\documentclass[letterpaper, 10 pt, conference]{ieeeconf}  
\IEEEoverridecommandlockouts 
\usepackage{amsmath} 
\usepackage{amssymb}  
\usepackage{amsthm}

\usepackage[short, c3, nocomma]{optidef}
\usepackage{cite}
\usepackage{graphicx}
\graphicspath{{images/}}

\usepackage{blindtext}
\usepackage{subfiles} 

\usepackage[dvipsnames]{xcolor}
\usepackage{blkarray, bigstrut, booktabs}

\usepackage{comment}

\newtheorem{thm}{Theorem}
\newtheorem{lem}{Lemma}
\newtheorem{prop}{Proposition}

\newtheorem{assumption}{Assumption}
\theoremstyle{definition}

\newtheorem*{remark}{Remark}
\newtheorem{problem}{Problem}

\theoremstyle{definition}

\newcommand{\support}{\sigma}
\newcommand{\cost}{\mu}
\newcommand{\Hp}{\mathcal{H}^+}
\newcommand{\Hm}{\mathcal{H}^-}
\newcommand{\A}{\mathcal{A}}
\newcommand{\B}{\mathcal{B}}
\newcommand{\Veh}{\mathcal{V}}
\newcommand{\Obs}{\mathcal{O}}
\newcommand{\pen}{\text{pen}}
\newcommand{\sd}{\textup{sd}}
\newcommand{\sv}{\textup{sv}}

\newcommand{\Rot}{R}

\newcommand{\fd}{f_{\Delta T}}

\title{A Differentiable Signed Distance Representation for Continuous Collision Avoidance in Optimization-Based Motion Planning}
\author{James Guthrie\thanks{J. Guthrie is with the Department of Electrical and Computer Engineering, 3400 N. Charles Street, Johns Hopkins University, Baltimore, MD 21218, USA. Email: {\tt\small guthriejd1@gmail.com} }}
\date{ }

\begin{document}

\maketitle
\thispagestyle{empty}
\pagestyle{empty}

\begin{abstract}
    This paper proposes a new set of conditions for exactly representing collision avoidance constraints within optimization-based motion planning algorithms. The conditions are continuously differentiable and therefore suitable for use with standard nonlinear optimization solvers. The method represents convex shapes using  a support function representation and is therefore quite general. For collision avoidance involving polyhedral or ellipsoidal shapes, the proposed method introduces fewer variables and constraints than existing approaches. Additionally the proposed method can be used to rigorously ensure continuous collision avoidance as the vehicle transitions between the discrete poses determined by the motion planning algorithm. Numerical examples demonstrate how this can be used to prevent problems of corner cutting and passing through obstacles which can occur when collision avoidance is only enforced at discrete time steps.
\end{abstract}
\section{Introduction}

Motion planning is a core component of the autonomy stack in robots, drones, and self-driving cars. While many approaches exist, motion planning algorithms based on optimization methods are popular for the ease with which they can represent a wide variety of objectives, constraints, and dynamic models. This approach relies on transcription methods such as direct multiple shooting  \cite{Bock1984} and direct collocation \cite{Biegler1984} which approximate a continuous-time optimal control problem as a discrete-time optimal control problem suitable to nonlinear programming (NLP) solvers.

Most NLP solvers utilize gradient and Hessian information and require that the objective and constraints imposed be twice continuously differentiable (i.e. belonging to the class of $\mathcal{C}^2$ functions) to guarantee convergence to a local minima. This presents a challenge for representing collision avoidance constraints between a vehicle occupying space $\Veh$ and an obstacle $\Obs$ as we generally lack smooth, closed-form representations of the condition $\Veh \cap \Obs = \emptyset$. 

One notable exception is when we can represent the indicator function of the Minkowski sum $\Veh \oplus (-\Obs)$ as a differentiable, closed-form expression. As $\Veh \cap \Obs = \emptyset \iff 0 \not\in \Veh \oplus (-\Obs)$, the latter provides a suitable constraint for ensuring collision avoidance. This is possible when both $\Veh$ and $\Obs$ are ellipsoids~\cite{Yan2015}. Alternatively, it is sometimes possible to closely approximate the Minkowski sum in closed-form using sum-of-squares optimization~\cite{Guthrie2022}. When available, closed-form Minkowski sums are appealing as they can be used to ensure collision avoidance without requiring additional variables be introduced into the problem.

A series of works have shown how collision avoidance conditions can be suitably represented within an NLP problem by introducing a set of differentiable conditions and auxiliary variables that collectively ensure $\Veh \cap \Obs = \emptyset$. All of these methods focus on specific classes of convex sets and then leverage various results from convex analysis which provide certificates that two sets do not intersect. In \cite{Patel2011} the authors utilize a polar set representation of polyhedrons to establish differentiable conditions for ensuring a point mass vehicle does not make contact with a polyhedral obstacle. In \cite{Gerdts2012} the authors leveraged Farkas' Lemma to arrive at conditions ensuring collision avoidance between a polyhedral robot and polyhedral obstacle. In \cite{Zhang2021} the authors utilized the dual formulation of distance calculations as given in \cite{Boyd2004} to ensure a minimum signed distance (a generalization of collision avoidance) between convex objects modeled as the intersection of linear and second-order cone constraints. All of these works require introducing additional variables and constraints into the problem. Generally the number of variables is proportional to the complexity of the geometry being represented. The resulting growth in problem size can quickly become burdensome. 

Beyond the challenge of computational complexity, all of these methods only address collision avoidance at discrete time instances arising from the transcription method utilized. The solver may exploit this discrete approximation of a continuous-time problem and return solutions which cut corners or pass through thin walls in an attempt to minimize the objective. In computational geometry this is a well-studied problem known as ``tunneling" as it can occur when a fast-moving bullet in a video game passes through thin walls.  Continuous collision detection refers to the class of algorithms in computational geometry which ensure robust collision checking at all time instances, not just discrete time points (e.g. between frame updates in a video game). These methods often rely on various approximations of the swept volume \cite{Choi2009}. 

Ensuring continuous collision detection within optimization-based motion planners is an open issue. In \cite{Dueri2017}, the authors present an exact approach for a specific class of dynamic models controlling point-mass vehicles navigating circles and cylinders. In \cite{Schulman2014} the authors develop a trajectory optimization algorithm that approximately ensures continuous collision detection for polyhedral robots navigating polyhedral obstacles. The method utilizes a linear approximation of the non-differentiable signed distance function. At points of non-differentiability, the resulting gradient information is inaccurate making the method ill-suited for use with standard NLP solvers which expect exact gradients. Instead the authors provide a custom solver based on sequential convex optimization. Beyond this, to the author's knowledge, no other methods exist for rigorously addressing continuous collision avoidance within optimization-based motion planners. Instead, various heuristic fixes are generally utilized. The most common is to inflate obstacles along with introducing velocity constraints on the vehicle to prevent it from passing through an obstacle in one time step \cite{LaValle2006}. However, this artificially reduces the configuration space of the problem, making tight maneuvering infeasible. Additionally it typically requires a smaller time step, leading to more decision variables in the transcription method and therefore larger (slower) optimization problems.

\subsection{Contributions}
In this work, we propose a novel formulation of signed distance constraints for collision avoidance by deriving necessary and sufficient conditions related to the support function representation of convex sets. These conditions are continuously differentiable and can be utilized within standard optimization-based motion planning algorithms based on nonlinear programming. Compared to existing approaches \cite{Patel2011, Gerdts2012, Zhang2021}, our method introduces fewer variables and constraints leading to smaller nonlinear programs. Additionally our formulation allows us to represent sets given by the convex hull of other sets. We utilize this capability to develop sufficient conditions for ensuring continuous collision avoidance within an optimization-based motion planning algorithm. To our knowledge, this is the first method for rigorously ensuring continuous collision avoidance within optimization-based motion planners for arbitrary vehicle dynamics and full-dimensional (vice point mass) geometries. We demonstrate its use on an autonomous vehicle model performing tight maneuvering around obstacles in which a discrete collision avoidance approach fails. 

\section{Background}

\subsection{Notation}
Let $[n] := \{1,2,\hdots, n \}$. Let $\mathbb{S}^n_{++}$ denote the set of $n \times n$ positive definite matrices. Let $SO(n)$ denote the special-orthogonal group in dimension $n$. Let $\|c\| := \|c\|_2$, denote the Euclidean norm of $c \in \mathbb{R}^n$. Let $B_r := \{ x \, | \, \|x\|_2 \leq r \}$. Given $\mathcal{A} \subset \mathbb{R}^n$, $\Rot \in \mathbb{R}^{n \times n}$, and $v \in \mathbb{R}^n$, let $\Rot\mathcal{A} + v := \{ \Rot x + v \, | \, x \in \mathcal{A} \}$.

\subsection{Signed Distance}
Let $\Veh,\Obs \subset \mathbb{R}^n$ be compact sets. The distance between the two objects is
\begin{equation}
    \text{dist}(\Veh,\Obs) := \underset{v}{\text{min}} \{\|v\| \,| \, (\Veh + v) \cap \Obs \neq \emptyset \}.
\end{equation}
If both $\Veh$ and $\Obs$ are convex, the distance can be calculated using convex optimization.
The penetration depth is
\begin{equation}
    \text{pen}(\Veh,\Obs) := \underset{v}{\text{min}} \{\|v\| \,| \, (\Veh + v) \cap (\Obs \setminus \partial\Obs) = \emptyset \}.
\end{equation}
The penetration depth is the minimum translation needed for $\Veh$ to not touch the interior of $\Obs$. Unlike distance calculations involving convex sets, calculating the penetration depth of two convex sets is a non-convex optimization problem with possibly multiple local minima. 
The signed distance combines the notions of distance and penetration and is given by
\begin{equation}
    \text{sd}(\Veh,\Obs) := \text{dist}(\Veh,\Obs) - \text{pen}(\Veh,\Obs).
\end{equation}
A positive signed distance indicates two objects are separated, a negative signed distance indicates they overlap, and a signed distance of zero indicates their boundaries touch. 

\subsection{Support and Cost Functions}
Let $\mathcal{S} \subseteq \mathbb{R}^n$ and $c \in \mathbb{R}^n \setminus 0$. The support function of $\mathcal{S}$ is
\begin{equation}
    \sigma_{\mathcal{S}}(c) := \underset{x \in \mathcal{S}}{\textup{sup}} \, c^Tx
\end{equation}
We will find it convenient to define the following function, which we refer to as the cost function of $\mathcal{S}$:
\begin{equation}
    \mu_{\mathcal{S}}(c) := \underset{x \in \mathcal{S}}{\textup{inf}} \, c^Tx
\end{equation}
These are related by $\mu_{\mathcal{S}}(c) = -\sigma_{\mathcal{S}}(-c)$.

\begin{prop}\label{supp_cost_properties}
Let $\A, \B \subset \mathbb{R}^n$ be convex sets. Let $t, c \in \mathbb{R}^n, R \in \mathbb{R}^{m \times n}, k \in \mathbb{R}_{\geq 0}$. The support and cost functions satisfy the following properties \cite{Schneider2013}: 
\begin{itemize}
    \item Scaling: 
    \begin{align*}
        \support_{k\A}(c) = k\support_{\A}(c) ,\;
        \cost_{k\A}(c) = k\cost_{\A}(c)
    \end{align*}
    \item Linear Transformation:
    \begin{align*}
        \support_{R\A}(c) = \support_{\A}(R^Tc) ,\;
        \cost_{R\A}(c) = \cost_{\A}(R^Tc)
    \end{align*}
    \item Translation: 
        \begin{align*}
            \support_{\A + v}(c) &= \support_{\A}(c) + c^Tv \\
            \cost_{\A + v}(c) &= \cost_{\A}(c) + c^Tv
    \end{align*}
    \item Minkowski Sum:
        \begin{align*}
            \support_{\A \oplus \B}(c) &= \support_\A(c) + \support_\B(c) \\
            \cost_{\A \oplus \B}(c) &= \cost_\A(c) + \cost_\B(c)
    \end{align*}
    \item Convex Hull:
        \begin{align*}
            \support_{\textup{co}(\{\A,\B\})}(c) &= \textup{sup}\{\support_\A(c), \support_\B(c)\} \\
            \cost_{\textup{co}(\{\A,\B\})}(c) &= \textup{inf}\{\cost_\A(c), \cost_\B(c)\}.
    \end{align*}
\end{itemize}
\end{prop}

The following lemmas will prove useful in relating the signed distance between two sets to their respective support and cost functions.
\begin{lem} \label{lem:sd_halfplane}
Let $\alpha, \beta \in \mathbb{R}$ and $c \in \mathbb{R}^n$, $\|c\| = 1$. Given halfspaces $\Hp= \{ x \, | \, c^Tx \geq \alpha \}$, $\Hm = \{ x \, | \, c^Tx \leq \beta \}$, then
\begin{equation}
    \sd(\Hp,\Hm) = \alpha - \beta.
\end{equation}
\end{lem}

\newcommand{\C}{\Veh}
\newcommand{\D}{\Obs}
\begin{lem}\label{lem:sd_subset}
Given $\C \subseteq \C^+ \subseteq \mathbb{R}^n$, 
$\D \subseteq \D^+ \subseteq \mathbb{R}^n$, then
\begin{equation}
\sd(\C, \D) \geq \sd(\C^+, \D^+).
\end{equation}

\end{lem}

\section{Problem Description}
\subsection{Vehicle Dynamics}
Consider a continuous-time model of a vehicle with state $x \in \mathbb{R}^{n_x}$, control input $u \in \mathbb{R}^{n_u}$, and dynamics $f: \mathbb{R}^{n_x} \times \mathbb{R}^{n_u} \rightarrow \mathbb{R}^{n_x}$ satisfying
\begin{equation} \label{eqn:xdot}
    \dot{x} = f(x,u).
\end{equation}
 In numerical optimal control, it is common to approximate continuous-time dynamics with a discrete-time model. The discrete model is obtained by applying a numerical integration method (e.g. Euler, Runge-Kutta) to the continuous-time dynamics over a fixed time interval $\Delta T$. The state and control values are then represented at indices $k \in \mathbb{Z}_+$ corresponding to their values in continuous time at $t_k = k\Delta T$. Let $x_k$ and $u_k$ denote the state and control respectively at time $t_k$.
 The value $u_k$ represents a constant control input applied for $t \in [t_k, t_{k+1})$.
 Let $\phi(x_i,\bar{u},t_i,t) := x_i + \int_{t_i}^{t}f(x(s),\bar{u})ds$ denote the solution of \eqref{eqn:xdot} at time $t \geq t_i$ with initial state $x_i$ and constant control input $u = \bar{u}$. 
 The resulting discrete-time model is given by\footnote{Throughout this work, we assume this relation holds exactly such that the discrete-time model has no integration error.}
\begin{equation} \label{eqn:xdot_discrete}
\begin{split}
   x_{k+1} &= \phi(x_k, u_k, t_k, t_k + \Delta T) \\
           :&= \fd(x_k, u_k).
\end{split}
\end{equation}
We refer to $\fd: \mathbb{R}^{n_x} \times \mathbb{R}^{n_u} \rightarrow \mathbb{R}^{n_x}$ as the discrete-time model of \eqref{eqn:xdot} with step-size $\Delta T$.

\subsection{Vehicle Geometry}
Let $\A \subset \mathbb{R}^n$ be a compact convex set describing the shape of the vehicle with dynamics \eqref{eqn:xdot}. Let 
\begin{equation}\label{eqn:geo_veh}
    \Veh(x) := R(x)\A + p(x)
\end{equation}
denote the space occupied by the vehicle where $R: \mathbb{R}^{n_x} \rightarrow SO(n), p: \mathbb{R}^{n_x} \rightarrow \mathbb{R}^n$ define the rotation and translation respectively. We refer to $\Veh(x)$ as the state-dependent geometry of the vehicle.
The swept volume is defined as the total space occupied (temporarily) by the vehicle over a time interval $[t_i, t_f]$:
\begin{equation}\label{eqn:sv_continuous}
    \textup{sv}_{\Veh, f}(x_i, \bar{u}, t_i, t_f) := \bigcup_{t \in [t_i, t_f]} \Veh(\phi(x_i, \bar{u}, t_i, t)).
\end{equation}

If the vehicle only undergoes linear translation the swept volume is the convex hull of the start and end poses. 
\begin{lem} \label{lem:swept_linear_dynamic}
Let the vehicle have continuous-time dynamics \eqref{eqn:xdot} with associated geometry \eqref{eqn:geo_veh}. Given initial state $x_i$ and control input $\bar{u}$, let $x(t) := \phi(x_i, \bar{u}, t_i, t)$ denote the resulting state trajectory for $t \in [t_i, t_f]$. Let $x_f := \phi(x_i, \bar{u}, t_i, t_f)$.
 Assume $R(x(t)) = R(x_i) \, \forall \, t \in [t_i, t_f]$. Assume $p(x(t)) = (1-\xi(t))p(x_i) + \xi(t)p(x_f)  \, \forall \, t \in [t_i, t_f]$ where $\xi: [t_i, t_f] \rightarrow [0, 1]$ is a continuous function with $\xi(t_i) = 0, \xi(t_f) = 1$. Then
\begin{equation} \label{eqn:sv_hull_cont}
    \textup{sv}_{\Veh, f}(x_i, \bar{u}, t_i, t_f) = \textup{co}(\{\Veh(x_i), \Veh(x_f)\}).
\end{equation}
\end{lem}

If the vehicle undergoes rotation or nonlinear translation, the resulting swept volume is, in general, non-convex. Further, we cannot determine the swept volume solely from the start and end poses. This presents a challenge for representing the swept volume within a numerical optimal control problem which only models the vehicle pose at discrete time steps. We assume the existence of a function that allows us to outer approximate the swept volume given the start and end poses and control input applied.

\begin{assumption}[Swept Volume of Vehicle]\label{assm:sc_veh}
\newcommand{\Br}{B_{r(x_k,u_k)}}
Let the vehicle have continuous-time dynamics \eqref{eqn:xdot}, discrete-time dynamics \eqref{eqn:xdot_discrete} and associated geometry \eqref{eqn:geo_veh}. Let the swept volume be given by \eqref{eqn:sv_continuous}. Assume there exists a $\mathcal{C}^2$ function $r: \mathbb{R}^{n_x} \times \mathbb{R}^{n_u} \rightarrow \mathbb{R}_{\geq 0}$ satisfying 
\begin{align}
    \textup{sv}_{\Veh, f}(x_k, u_k, t_k, t_{k+1})  \subseteq \textup{co}(\{\Veh(x_k), \Veh(x_{k+1})\}) \oplus B_{r(x_k,u_k)} 
\end{align}

where 
\begin{equation}
   B_{r(x_k,u_k)} := \{ y \in \mathbb{R}^n \, | \, \|y\| \leq r(x_k,u_k)\}.
\end{equation}

\end{assumption}

\begin{figure}
    \vspace{3mm}
    \centering
    \includegraphics[width=0.47\textwidth]{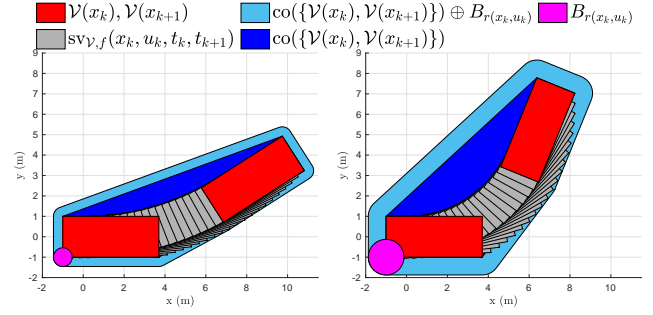}
    \caption{Swept volume of vehicle. As the vehicle turns more, the swept volume deviates more from the convex hull of the start and end poses.}
    \label{fig:swept_volume_outer}
\end{figure}

Figure \ref{fig:swept_volume_outer} visualizes this outer approximation. The ball $B_{r(x_k,u_k)}$ accounts for the amount by which the convex hull underapproximates the true swept volume. By making the ball's radius a function of the vehicle state and input, we can minimize the extent to which we overapproximate the swept volume. For example, when the vehicle is moving in a straight line, ideally we would have $r(x_k,u_k) = 0$.  

\subsection{Obstacle Geometry}
Let $\B \subset \mathbb{R}^n$ be a closed convex set describing the shape of an obstacle. Let
\begin{equation}\label{eqn:geo_obs_continuous}
    \Obs(t) := S(t)\B + d(t)
\end{equation}
denote the space occupied by the obstacle at time $t$ where 
$S: \mathbb{R} \rightarrow SO(n), d: \mathbb{R} \rightarrow \mathbb{R}^n$ define the rotation and translation respectively.\footnote{Our notation is chosen to support moving obstacles. For stationary objects we replace $S(t)$ and $d(t)$ with constants.} We refer to $\Obs(t)$ as the time-dependent geometry of the obstacle. The swept volume of the obstacle is defined as the total space occupied over a time interval $[t_i, t_f]$:
\begin{equation}\label{eqn:sv_obs_continuous}
    \textup{sv}_{\Obs}(t_i, t_f) := \bigcup_{t \in [t_i, t_f]}\Obs(t).
\end{equation}
In our setting, we are only given the obstacle's pose at $t_k := k\Delta T, k \in \mathbb{Z}_+$, where $\Delta T$ is the time step-size. For convenience, let $S_k := S(t_k), d_k := d(t_k)$ and $\Obs_k := \Obs(t_k)$ such that the obstacle's pose at time index $k$ is
\begin{equation}\label{eqn:geo_obs}
    \mathcal{O}_k = S_k\mathcal{B} + d_k.
\end{equation}
  We assume that the obstacle's swept volume belongs to the convex hull of the start and end poses inflated by ball $B_{w_k}$.
\begin{assumption}[Swept Volume of Obstacle]\label{assm:sv_obs}
Let the obstacle have continuous-time geometry \eqref{eqn:geo_obs_continuous} and discrete-time geometry \eqref{eqn:geo_obs}. Let the swept volume be given by \eqref{eqn:sv_obs_continuous}. Let $w_k \in \mathbb{R}_{\geq 0}$ satisfy
\begin{equation}
    \sv_\Obs(t_k, t_{k+1}) \subseteq \textup{co}(\{\Obs_k, \Obs_{k+1}\}) \oplus B_{w_k}.
\end{equation}
\end{assumption}
\subsection{Optimization-Based Motion Planning}
Consider generating a motion plan over a time horizon $t \in [0, \, T_f]$ for a vehicle with continuous dynamics and geometry given by \eqref{eqn:xdot} and \eqref{eqn:geo_veh} respectively. We use a discrete representation of the dynamics as given by \eqref{eqn:xdot_discrete} with $N \Delta T = T_f$ and $k \in \{0,\hdots,N\}$ for some $N \in \mathbb{Z}_+$. The vehicle starts at state $x_S$ and ends at final state $x_F$. The vehicle must maintain a minimum signed distance of $\gamma$ to an obstacle with geometry given by \eqref{eqn:geo_obs_continuous}.\footnote{We consider a single obstacle to minimize notational clutter. This is without loss of generality as the conditions developed can be repeatedly applied to address the case of multiple obstacles.} Let $X := [x_0^T,\hdots, x_{N}^T]^T$ and $U := [u_0^T,\hdots,u_{N-1}]^T$ denote the vector of all states and controls respectively. We seek to minimize an objective $l(X,U)$ where $l: X \times U \rightarrow \mathbb{R}$. The vehicle is subject to constraints $h(X,U) \leq 0$ where $h: X \times U \rightarrow \mathbb{R}^{n_h}$ and the inequality is interpreted element-wise. We assume that $l(X,U)$ and $h(X,U)$ are $\mathcal{C}^2$ functions. The resulting optimization problem is given by

\begin{mini}|s|
{X,U}{l(X,U)}{\label{opti:trajopt}}{}
\addConstraint{x_0 = x_S, \quad}{x_N = x_F,}{}
\addConstraint{x_{k+1}}{=\fd(x_k,u_k),}{\,k = 0,\hdots,N-1,}
\addConstraint{h(X,U)}{\leq 0,}{}
\addConstraint{\sd(\Veh, \Obs)}{\geq \gamma}{}.
\end{mini}

The signed distance function is in general, non-smooth and lacks a closed-form representation. We focus on establishing $\mathcal{C}^2$ conditions that can equivalently represent the signed distance constraints. We first address the case in which the signed distance constraint is imposed at discrete time steps.
\begin{problem}[Discrete Collision Avoidance]\label{prob:dca}
Consider the motion planning problem given by \eqref{opti:trajopt}. Find a set of $\mathcal{C}^2$ constraints that ensure a minimum signed distance of $\gamma$ at discrete time step $t_k$ using the vehicle state $x_k$ and additional variables $y \in \mathbb{R}^{n_y}$:
\begin{equation}
    \sd(\Veh(x_k),\Obs_k) \geq \gamma \iff \exists \, x_k,y \,|\, h(x_k,y) \leq 0
\end{equation}
where $h: \mathbb{R}^{n_x} \times \mathbb{R}^{n_y} \rightarrow \mathbb{R}^{n_c}$ is $\mathcal{C}^2$ and the inequality constraint is interpreted element-wise.
\end{problem}

Problem \ref{prob:dca} only ensures the signed distance constraint is satisfied at time $t_k$. To ensure the continuous-time trajectory satisfies the signed distance constraint, we evaluate the signed distance using the swept volumes of the vehicle and obstacle over the interval $t \in [t_k, t_{k+1}]$. 

\begin{problem}[Continuous Collision Avoidance]\label{prob:cca}
Consider the motion planning problem given by \eqref{opti:trajopt}. Let the vehicle and obstacle swept volumes satisfy Assumptions \ref{assm:sc_veh} and \ref{assm:sv_obs}. Find a set of $\mathcal{C}^2$ constraints that ensure a minimum signed distance of $\gamma$ for $t \in [t_k, t_{k+1}]$ using the vehicle state $x_k$, input $u_k$ and additional variables $y \in \mathbb{R}^{n_y}$:
\begin{equation}
\begin{split}
    \sd ( \textup{sv}_{\Veh, f}(x_k, u_k, t_k, t_{k+1}), \sv_{\Obs}(t_k, t_{k+1})) \geq \gamma 
    \\ \Leftarrow  
    \exists \, x_k,u_k,y \,|\, h(x_k,u_k,y) \leq 0
\end{split}
\end{equation}
where $h: \mathbb{R}^{n_x} \times \mathbb{R}^{n_u} \times \mathbb{R}^{n_y} \rightarrow \mathbb{R}^{n_c}$ is $\mathcal{C}^2$ and the inequality constraint is interpreted element-wise.
\end{problem}

\section{A Differentiable Signed Distance Representation}
\newcommand{\costA}{\cost_\A(R(x_k)^Tc)}
\newcommand{\supportB}{\support_\B(S_k^Tc)}

We now develop differentiable representations of the signed distance constraints. We focus on establishing this representation for one time step $t_k$ or time interval $[t_k, t_{k+1}]$. This is without loss of generality as these conditions can be repeatedly applied to address multiple time steps or intervals.

\subsection{Discrete Collision Avoidance}
\newcommand{\C}{\mathcal{C}}
\newcommand{\D}{\mathcal{D}}
The following lemmas relate the signed distance between two convex sets $\C$ and $\D$ to their cost and support function respectively evaluated for a given vector $c$.

\begin{lem} \label{lem:sd_cost_minus_support}
Given $\C, \D \subseteq \mathbb{R}^n, c \in \mathbb{R}^n, \|c\| = 1$ then
\begin{equation}
    \sd(\C,\D) \geq \cost_\C(c) - \support_\D(c).
\end{equation}
\end{lem}

\begin{lem} \label{lem:sd_c_exists}
Let $\C, \D \subseteq \mathbb{R}^n$ be closed convex sets. Let $\C$ and/or $\D$ be bounded. Then there exists $c \in \mathbb{R}^n, \|c\| = 1$ such that
\begin{equation}
    \sd(\C,\D) = \cost_\C(c) - \support_\D(c).
\end{equation}
\end{lem}

Lemma \ref{lem:sd_cost_minus_support} suggests a simple method for representing signed distance constraints within a nonlinear program. We introduce a decision variable $c \in \mathbb{R}^n, \|c\| = 1$ along with constraints that make $c$ define a certificate that $\sd (\Veh(x_k), \Obs_k) \geq \gamma$. Lemma \ref{lem:sd_c_exists} guarantees that such a certificate exists.

We will find it convenient to rewrite the cost and support of $\Veh(x_k)$ and $\Obs_k$ in terms of the base shape $\A$ and $\B$. Using the properties listed in Proposition \ref{supp_cost_properties} yields
\begin{equation}
\begin{split}
    &\cost_{\Veh(x_k)}(c) - \support_{\Obs_k}(c) \\
    &= \cost_{R(x_k)\A + p(x_k)}(c) - \support_{S_k\B + d_k}(c) \\
    &= (\cost_{R(x_k)\A}(c) + c^Tp(x_k)) - (\support_{S_k\B}(c) + c^Td_k) \\  
    &= \cost_{\A}(R(x_k)^Tc) - \support_{\B}(S_k^Tc)  + c^T(p(x_k)-d_k).
\end{split}    
\end{equation}

\begin{lem}\label{smooth_sd_1}
Let the vehicle geometry $\Veh(x_k)$ be given by \eqref{eqn:geo_veh}. Let the obstacle geometry $\Obs_k$ be given by \eqref{eqn:geo_obs}.
Then $\sd(\Veh(x_k),\Obs_k) \geq \gamma$ if and only if there exists $c \in \mathbb{R}^n$ satisfying:
\begin{align}
    \gamma &\leq \cost_{\A}(R(x_k)^Tc) - \support_{\B}(S_k^Tc) + c^T(p(x_k) -d_k), \label{eqn:lhs_1}\\
    1 &= \|c\|. \label{eqn:con_norm_equality}
\end{align}
\end{lem}
\begin{proof}
$\Leftarrow$: From Lemma \ref{lem:sd_cost_minus_support}, $\cost_{\Veh(x_k)}(c) - \support_{\Obs_k}(c) \geq \gamma \implies \sd(\Veh(x_k),\Obs_k) \geq \gamma$.
$\Rightarrow$: Let $\sd(\Veh(x_k),\Obs_k) = \eta$ where $\eta \geq \gamma$. From Lemma \ref{lem:sd_c_exists}, there exists $c \in \mathbb{R}^n, \|c\| = 1$ such that $\cost_{\Veh(x_k)}(c) - \support_{\Obs_k}(c) = \eta$.
\end{proof}

\newcommand{\scalec}{\frac{c}{\|c\|}}
\newcommand{\scalek}{\frac{1}{\|c\|}}
\begin{remark}
If $\gamma > 0$ we can relax \eqref{eqn:con_norm_equality} to the convex constraint $\|c\| \leq 1$. To see this, first note that $c = 0$ cannot satisfy \eqref{eqn:lhs_1} for $\gamma > 0$ as the right-hand side will evaluate to zero. Now consider a solution $c$ in which $0 < \|c\| < 1$. Multiplying \eqref{eqn:lhs_1} by $\scalek$ we obtain

\begin{align*}
    \scalek \gamma &\leq \scalek(\cost_\A(R(x_k)^Tc) - \support_B(S_k^Tc) + c^T(p(x_k) - d_k) \\
    &= \cost_A(R(x_k)^T\scalec) \! - \support_B(S_k^T\scalec) + \frac{c^T}{\|c\|}(p(x_k)\! -\! d_k).
\end{align*} 
Let $\tilde{c} = \frac{c}{\|c\|}$. From Lemma \ref{lem:sd_cost_minus_support}, $\tilde{c}$ provides a certificate that $\sd(\Veh(x_k),\Obs_k) \geq \scalek \gamma > \gamma$.
\end{remark}
 
Lemma \ref{smooth_sd_1} provides a differentiable representation of signed distance constraints in the case that $\cost_\A(c), \support_\B(c)$ are given by $\mathcal{C}^2$ functions. Points and ellipsoids satisfy this condition. We now leverage the convex hull property of the cost and support functions to represent shapes defined by the convex hull of multiple convex sets.
\begin{thm}\label{smooth_sd_2}
Let the vehicle geometry $\Veh(x_k)$ be given by \eqref{eqn:geo_veh}. Let the obstacle geometry $\Obs_k$ be given by \eqref{eqn:geo_obs}. Let $\A = \textup{co}(\{\A^{(i)}, i \in [n_\A]\})$ and $\B = \textup{co}(\{\B^{(j)}, j \in [n_\B]\})$ where each $\A^{(i)}, \B^{(j)}$ is a convex set. Then $\sd(\Veh(x_k),\Obs_k) \geq \gamma$ if and only if there exists $c \in \mathbb{R}^n, \alpha, \beta \in \mathbb{R}$ satisfying
\begin{align*}
    \alpha &\leq \cost_{\A^{(i)}}(R(x_k)^Tc), &i \in [n_\A], \\
    \beta &\geq \support_{\B^{(j)}}(S_k^Tc),  &j \in [n_\B], \\
    \gamma &\leq \alpha - \beta + c^T(p(x_k) - d_k), \\
    1 &= \|c\|.
\end{align*}
\end{thm}

\newcommand{\costAi}{\cost_{\A^{(i)}}(R(x_k)^Tc)}
\newcommand{\supportBj}{\support_{\B^{(j)}}(S_k^Tc)}
\begin{proof}
Note that $\alpha \leq \costAi, i \in [n_\A] \implies \alpha \leq \cost_\A(R(x_k)^Tc)$ by the convex hull property of the cost function. Similarly, $\beta \geq \supportBj, j \in [n_\B] \implies \beta \geq \support_\B(S_k^Tc)$. The remainder of the proof follows the same arguments as Lemma \ref{smooth_sd_1}.
\end{proof}

\begin{remark}
 If $n_\A = 1$ in Theorem \ref{smooth_sd_2} then we can set $\alpha = \costA$ without loss of generality. The variable $\alpha$ can be eliminated (replaced with $\costA$). Similarly, if $n_\B = 1$ we can eliminate the variable $\beta$. If $n_\A = 1$ and $n_\B = 1$, Theorem \ref{smooth_sd_2} reduces to Lemma \ref{smooth_sd_1}. 
\end{remark}

\subsection{Examples}
We now apply Theorem \ref{smooth_sd_2} to obtain collision avoidance conditions for polyhedral and ellipsoidal shapes. In doing so, we will see that this formulation introduces fewer variables and constraints than the duality-based formulation of \cite{Zhang2021}. This can be beneficial for reducing the computational complexity of the nonlinear program. Although our examples are limited to cases in which the vehicle and obstacle shape are of the same class, it is straight-forward to extend these results to cases in which different classes are present (e.g. polyhedral vehicle and ellipsoidal obstacle).

\subsubsection{Polyhedrons}
Consider the case in which both the vehicle shape and obstacle shape are convex, compact polyhedrons in $\mathbb{R}^n$ with $n_\A$ and $n_\B$ vertices respectively:
\begin{align*}
    \A &= \textup{co}(\{ a_i \in \mathbb{R}^n, i \in [n_\A]\}), \\
    \B &= \textup{co}(\{ b_j \in \mathbb{R}^n, j \in [n_\B]\}).
\end{align*}
 Note that for a single point $q \in \mathbb{R}^n$ we have $\cost_q(c) = \support_q(c) = c^Tq$. Using Theorem \ref{smooth_sd_2} we obtain conditions to ensure a minimum signed distance of $\gamma$ between two polyhedrons:
\begin{align*}
    \alpha &\leq c^TR(x_k)a_i, \, &i \in [n_\A] \\
    \beta &\geq c^TS_kb_j, \, &j \in [n_\B] \\
    \gamma &\leq \alpha - \beta + c^T(p(x_k) - d_k), \\
    1 &= \| c \|
\end{align*}
\begin{remark}
We contrast this with the method of \cite{Zhang2021} which assumes a halfspace-representation (vice vertex representation) of a compact polyhedron. Let $m_\A, m_\B$ be the number of linear constraints necessary to describe $\A, \B$ respectively. The dual approach introduces $(m_\A + m_\B)$ variables and $(2 + n + m_\A + m_\B)$ constraints to represent the signed distance constraint. Note we must have $m_\A, m_\B \geq n+1$ for $\A,\B$ to be compact with non-empty interior. Our formulation introduces $(2+n)$ variables and ($2 + n_\A + n_\B)$ constraints. For the case in which $n_\A = m_\A, n_\B = m_\B$, our method introduces fewer variables and fewer constraints.\footnote{A similar remark applies to the method of \cite{Gerdts2012} which uses Farkas' Lemma. This requires introducing $(m_\A + m_\B)$ variables and $(2(m_\A + m_\B) + 1)$ constraints.}
\end{remark}

\subsubsection{Ellipsoids}
Let the vehicle shape and obstacle shape be ellipsoids given by matrices $P,Q \in \mathbb{S}^n_{++}$:
\begin{align*}
    \A = \{ x \in \mathbb{R}^n \,|\, x^TP^{-1}x \leq 1 \}, \\
    \B = \{ x \in \mathbb{R}^n \,|\, x^TQ^{-1}x \leq 1 \}.
\end{align*}
Recall that ellipsoids have a closed-form cost and support function given by $\cost_\A(c) = -\sqrt{c^TPc}$ and $\support_\B(c) = \sqrt{c^TQc}$. 
As $n_\A = 1$ and $n_\B = 1$, Theorem \ref{smooth_sd_2} reduces to Lemma \ref{smooth_sd_1} yielding:
\begin{align*}
    \gamma &\leq -\sqrt{c^TR(x_k)PR(x_k)^Tc} - \sqrt{c^TS_kQS_k^Tc} \\ 
    & \; \; \; \; + c^T(p(x_k) - d_k), \\
    1 &= \| c \|.
\end{align*}

\begin{remark}
 The dual formulation in \cite{Zhang2021} uses second-order cone constraints to represent ellipsoids. Each second-order cone constraint introduces a dual variable pair $\lambda \in \mathbb{R}, u \in \mathbb{R}^n$ and the constraint $\lambda \geq \|u\|$. In total the dual formulation introduces $2(n+1)$ variables and $(4+n)$ constraints. Our formulation introduces $n$ variables and two constraints.
\end{remark}
\begin{remark}
The support function of an ellipsoid involves the square root, which is not differentiable at the origin. Given $P \succ 0$, the argument only evaluates to zero for $c = 0$ which cannot be a solution. However, we may experience issues if the solver is initialized with $c = 0$. We can add a small smoothing term $\epsilon > 0$ to address this case. Noting that $-\sqrt{c^TPc + \epsilon}  < -\sqrt{c^TPc}$ it is seen that this modification is conservative in that satisfying the conditions of Lemma \ref{smooth_sd_1} means $\cost_{\Veh(x_k)}(c) - \support_{\Obs_k}(c) > \gamma$. The signed distance constraint is then strictly satisfied.
\end{remark}

\subsection{Continuous Collision Avoidance}
Theorem \ref{smooth_sd_2} provides differentiable conditions for representing signed distance constraints between a vehicle and obstacle at discrete time steps $t_k, k \in \mathbb{Z}_+$. As the vehicle and obstacle transition between these discrete poses, the signed distance constraint may not be satisfied. This can be resolved by enforcing signed distance constraints with respect to the swept volume of the vehicle and obstacle over the time interval $t \in [t_k, t_{k+1}]$. 

Assumptions \ref{assm:sc_veh} and \ref{assm:sv_obs} define outer approximations of the swept volume of the vehicle and obstacle respectively. These approximations utilize the convex hull and Minkowski sum operators. To account for the Minkowski sum operator we will make use of the following lemma.
\begin{lem}\label{lem:sd_ball}
Let $\C,\D \subset \mathbb{R}^n$ be closed convex sets. Let $\C$ and/or $\D$ be bounded. Let $r_\C, r_\D \in \mathbb{R}_{\geq 0}$. Then
\begin{equation}
    \sd(\C \oplus B_{r_\C},\D \oplus B_{r_\D}) = \sd(\C,\D) - r_\C - r_\D.
\end{equation}
\end{lem}
The following relation results from applying Lemma \ref{lem:sd_subset} with Assumptions \ref{assm:sc_veh} and \ref{assm:sv_obs} followed by Lemma \ref{lem:sd_ball}:
\begin{equation}\label{eqn:suff_cond}
    \begin{split}
    \sd (&\sv_{\Veh,f}(x_k, u_k, t_k, t_{k+1}), \\
    &\sv_{\Obs}(t_k, t_{k+1})) \geq \gamma 
   \\ \Leftarrow
      \sd(&\textup{co}(\{\Veh(x_k), \Veh(x_{k+1})\}) \oplus B_{r(x_k,u_k)}, \\
    &\textup{co}(\{\Obs_k, \Obs_{k+1}\}) \oplus B_{w_k}) = \gamma
    \\\iff 
    \sd(&\textup{co}(\{\Veh(x_k), \Veh(x_{k+1})\}), \\
        &\textup{co}(\{\Obs_k, \Obs_{k+1}\})) = \gamma + r(x_k,u_k) + w_k.
    \end{split}
\end{equation}

From this relation, we can extend Theorem \ref{smooth_sd_2} to obtain sufficient conditions for continuous collision avoidance.
\begin{thm}\label{smooth_sd_3}
Let the vehicle dynamics and geometry satisfy Assumption \ref{assm:sc_veh}. Let the obstacle geometry satisfy Assumption \ref{assm:sv_obs}. 
Let $\A = \textup{co}(\{\A^{(i)}, i \in [n_\A]\})$ and $\B = \textup{co}(\{\B^{(j)}, j \in [n_\B]\})$ where each $\A^{(i)}, \B^{(j)}$ is a convex set. 
Then  $\sd (\sv_{\Veh,f}(x_k, u_k, t_k, t_{k+1}), \sv_\Obs(t_k, t_{k+1})) \geq \gamma$ if there exists $c \in \mathbb{R}^n, \alpha, \beta \in \mathbb{R}$ satisfying
\begin{align*}
    \alpha &\leq \cost_{\A^{(i)}}(R(x_k)^Tc), &i \in [n_\A] \\
    \alpha &\leq \cost_{\A^{(i)}}(R(x_{k+1})^Tc) + c^T(p(x_{k+1}) - p(x_k)), &i \in [n_\A] \\
    \beta &\geq \support_{\B^{(j)}}(S_k^Tc), &j \in [n_\B] \\
    \beta &\geq \support_{\B^{(j)}}(S_{k+1}^Tc) + c^T(d_{k+1} - d_k), &j \in [n_\B] \\
    \gamma &\leq \alpha - \beta + c^T(p(x_k) - d_k) - r(x_k,u_k) - w_k\\
    1 &= \|c\|.
\end{align*}
\end{thm}
\begin{proof}
 The stated conditions arise from applying the necessary and sufficient conditions of Theorem \ref{smooth_sd_2} to ensure a signed distance of $\gamma + r(x_k,u_k) + w_k$ between the sets $\textup{co}(\{\Veh(x_k), \Veh(x_{k+1})\})$ and $\textup{co}(\{\Obs_k, \Obs_{k+1}\})$. From \eqref{eqn:suff_cond}, this is sufficient for ensuring a signed distance of $\gamma$ between $\sv_{\Veh,f}(x_k, u_k, t_k, t_{k+1})$ and $\sv_\Obs(t_k, t_{k+1})$.
\end{proof}

\begin{remark}
Theorem \ref{smooth_sd_3} is only sufficient because we are outer-approximating non-convex swept volumes with convex sets. For example, in Figure \ref{fig:swept_volume_outer}, the right subplot shows an aggressive turn in which our outer approximation introduces noticeable conservatism. Here we are intentionally using a large integration step size $(\Delta T = 0.77s)$ to highlight this aspect. In practical applications, one can reduce the integration step size until this conservatism is acceptable. 
\end{remark}

\section{Examples}
We demonstrate our method using a car model navigating in $\mathbb{R}^2$. The vehicle state consists of positions $(p_x, p_y)$, orientation ($\psi$), velocity $(v)$, and steering angle ($\delta$). The inputs are acceleration $(a)$ and steering rate $(s)$. The parameter $L = 2.7$ is the wheelbase. The continuous-time dynamics are:
\begin{equation}
\begin{split}
    \dot{p}_x &= v\cos{\psi} \\
    \dot{p}_y &= v\sin{\psi} \\
    \dot{\psi} &= v\frac{\tan{\delta}}{L} \\
    \dot{v} &= a \\
    \dot{\delta} &= s
\end{split}
\end{equation}
The vehicle's shape is a polyhedron $\A = \textup{co}(\{ (\pm 2.5,\pm 1) \})$. The space occupied by the vehicle is given by
\begin{equation}
    \Veh(x) = R(x)\A + p(x)
\end{equation}
where
\begin{equation}
\begin{split}
    p(x) = \begin{bmatrix} p_x \\ p_y \end{bmatrix}, \,
    R(x) = 
    \begin{bmatrix}
    \cos{\psi} & -\sin{\psi} \\
    \sin{\psi} & \cos{\psi}
    \end{bmatrix}.
\end{split}
\end{equation}

We pose an optimal control problem in which the vehicle begins at $(p_x = 0, p_y = 25, \psi = 0)$ and must end at $(p_x = 100, p_y = 25, \psi = 0$) while minimizing the squared-norm of the control effort $l(X,U) = \|U\|_2^2$. We set the time horizon to 10s and use $N = 13$ steps, giving a discrete-time step of $\Delta T = \frac{10}{13}$. We use a 4th-order Runge-Kutta method to obtain the discrete dynamic model $x_{k+1} = \fd(x_k,u_k)$. We place a polyhedral obstacle $\Obs$ in the environment and solve \eqref{opti:trajopt} using both the discrete collision avoidance conditions (Theorem \ref{smooth_sd_2}) and the continuous collision avoidance conditions (Theorem \ref{smooth_sd_3}). 

\subsection{Swept Volume Approximation Model}
The continuous collision avoidance conditions require a $\mathcal{C}^2$ function $r: \mathbb{R}^{n_x} \times \mathbb{R}^{n_u} \rightarrow \mathbb{R}_{\geq 0}$ satisfying Assumption \ref{assm:sc_veh}. Although finding this function is not the focus of this work, we briefly sketch out a practical method for doing so. We first simulate the continuous dynamics over a time interval $t \in [0, \Delta T]$ where $\Delta T$ is the discrete time step used in the optimal control problem. We do this for various initial states $x_k^{(i)}$ and control inputs $u_k^{(i)}$ within expected ranges. For each sample $(x_k^{(i)}, u_k^{(i)})$ we compute the convex hull of the resulting swept volume.
We then compute the minimum radius $r^{(i)}$ such that 
$\textup{co}(\{\Veh(x_k^{(i)}), \Veh(x_{k+1}^{(i)})\}) \oplus B_{r^{(i)}} \supseteq \textup{co}(\textup{sv}_{\mathcal{V},f}(x_k,u_k,0,\Delta T))$. 
Finally we fit a non-negative function to the resulting data samples $\{x_k^{(i)}, u_k^{(i)}, r^{(i)} \}$. 
This can be done using sum-of-squares (SOS) optimization \cite{Parillo2000}. In our examples we utilized an 8th-order SOS polynomial $r(v_k,\delta_k)$ to represent the ball radius as a function of vehicle velocity and steering angle.

\subsection{Results}
\begin{figure*}
    \vspace{3mm}
    \centering
    \includegraphics[width=0.99\textwidth]{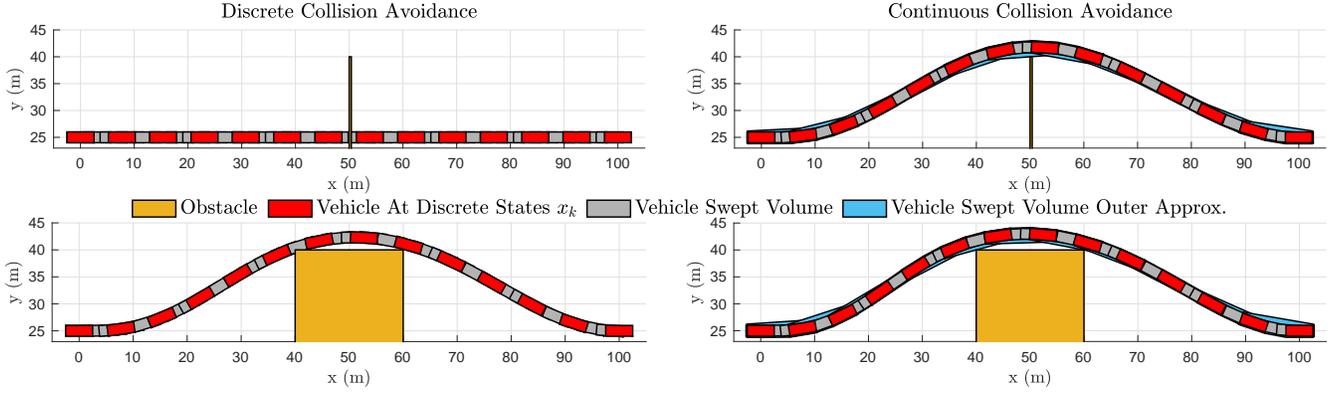}
    \caption{Autonomous car navigating obstacles. Discrete collision avoidance incorrectly passes through walls (upper left) and cuts corners (lower left). Continuous collision avoidance prevents these erroneous behaviors by checking collision with respect to an outer approximation of the swept volume.}
    \label{fig:dca_vs_cca}
\end{figure*}
\subsubsection{Thin Wall}
We first consider a thin wall and require $\sd(\Veh(x), \Obs) \geq 0$. When solving with the discrete collision avoidance conditions, the trajectory passes through the wall in order to minimize the control effort. We note that other methods for optimization-based collision avoidance such as \cite{Zhang2021} are susceptible to this behavior. This can occur even if the solver is initialized with a collision-free trajectory which navigates around obstacles. By utilizing the continuous collision avoidance conditions, the solver is prevented from exploiting the discrete approximation and returns a trajectory which successfully avoids the wall. Figure \ref{fig:dca_vs_cca} shows the results in the upper subplots. We note that the outer approximation of the swept volume is only slightly conservative compared to the true swept volume as shown by the blue borders. 

\subsubsection{Corner Cutting}
Another issue commonly faced by optimization-based motion planners is corner cutting. To demonstrate this, we replace the thin obstacle with a wide obstacle. Due to the velocity constraints on the vehicle, it is not possible for the discrete-time trajectory to ``jump over" the obstacle. Instead, the discrete collision avoidance constraints yield a trajectory that turns to avoid the obstacle. However, it cuts the corner at $(40,40)$ to minimize the necessary maneuvering. The continuous collision avoidance conditions again prevent this from happening. Figure \ref{fig:dca_vs_cca} shows the results in the lower subplots.

\subsection{Implementation Details}
All examples were solved on a MacBook Pro with a 2.6 GHz 6-Core Intel Core i7 CPU. IPOPT \cite{Wachter2006} with the MA27 linear solver was used to solve the nonlinear optimization problems with exact gradients and Hessians supplied by CasADi \cite{Andersson2019}. Supporting code is available at https://github.com/guthriejd1/cca.

\section{Conclusion}
A novel formulation of collision avoidance based on signed distance constraints was proposed for convex-shaped vehicles navigating convex obstacles. This formulation is continuously differentiable and therefore suitable for incorporation within optimization-based motion planning algorithms. For the important case of polyhedral and ellipsoidal shapes, this representation is more compact than existing formulations as it introduces fewer additional variables and constraints. Additionally, this formulation can be used to ensure the continuous-time trajectory satisfies the collision avoidance constraints despite being planned in a discrete setting. This provides a rigorous means of preventing ``tunneling" and corner-cutting which can occur when collision avoidance is only enforced at discrete time steps.

In future work we will document additional classes of convex sets that can be represented within this framework. We also will evaluate the numerical performance of this method against competing formulations for the discrete collision avoidance case. Lastly, we plan to explore more refined outer approximations of the swept volume to minimize conservatism.

\appendix
\section{Proofs}
\newcommand{\C}{\mathcal{C}}
\newcommand{\D}{\mathcal{D}}
\newcommand{\intD}{\D \setminus \partial \D}
\subsection{Proof of Lemma \ref{lem:sd_halfplane}}
\begin{proof}
We prove this for the $\alpha < \beta$ case. The $\alpha > \beta$ and $\alpha = \beta$ cases can be shown using similar arguments. Let $t = (\beta-\alpha)c$ and $x \in \Hp 
\implies c^T(x +t) \geq \alpha + (\beta - \alpha) = \beta
\implies x + t \not\in (\Hm \setminus \partial\Hm)
\implies \pen(\Hp,\Hm) \leq (\beta - \alpha)
\implies \sd(\Hp,\Hm) \geq (\alpha - \beta).
$
Assume $\sd(\Hp,\Hm) > \alpha - \beta 
\implies \pen(\Hp,\Hm) < \beta - \alpha
\implies \exists \, t \in \mathbb{R}^n, \|t\| < \beta-\alpha $ such that $
(\Hp + t) \cap (\Hm \setminus \partial\Hm) = \emptyset.$
Let $x \in \Hp$ satisfy $c^Tx  = \alpha
\implies c^T(x + t) = \alpha + c^Tt \leq \alpha + \|c\| \|t\| < \beta
\implies x+t \in (\Hm \setminus \partial\Hm)$ a contradiction. Therefore $\sd(\Hp,\Hm) = \alpha - \beta$. 

\end{proof}


\subsection{Proof of Lemma \ref{lem:sd_subset}}
\newcommand{\VehPlus}{\Veh^+}
\newcommand{\ObsPlus}{\Obs^+}
Let $\sd(\VehPlus,\ObsPlus) = -a \leq 0 \implies \pen(\VehPlus,\ObsPlus) = a \implies \exists \, t \in \mathbb{R}^n, \|t\| = a$ such that $(\VehPlus + t) \cap (\ObsPlus \setminus \partial \ObsPlus) = \emptyset \implies (\Veh + t) \cap (\Obs \setminus \partial \Obs) = \emptyset \implies \pen(\Veh,\Obs) \leq a \implies \sd(\Veh,\Obs) \geq -a$. The $\sd(\VehPlus,\ObsPlus) > 0$ case can be shown using similar arguments.

\subsection{Proof of Lemma \ref{lem:swept_linear_dynamic}}
\begin{proof}
Note that any $\Veh(x(t)) \in \textup{sv}_{\Veh, f}(x_i, \bar{u}, t_i, t_f)$ satisfies
\begin{equation}
    \begin{split}
          \Veh(x(t)) &= R(x_i)\A + (1-\xi(t))p(x_i) + \xi(t)p(x_f) \\
               &= (1-\xi(t))\Veh(x_i) + \xi(t)\Veh(x_f).
    \end{split}
\end{equation}
Given $\xi(t)$ is continuous with $\xi(t_i) = 0, \xi(t_f) = 1 \implies \forall \, \lambda \in [0, 1] \, \exists \, t \in [t_i, t_f] \, | \, \xi(t) = \lambda$. It follows that
\begin{equation}
\textup{sv}_{\Veh, f}(x_i, \bar{u}, t_i, t_f) = \textup{co}(\{\Veh(x_i), \Veh(x_f)\}).
\end{equation}
\end{proof}

\subsection{Proof of Lemma \ref{lem:sd_cost_minus_support}}

\begin{proof}
Define the following halfspaces:
\begin{align}
    \mathcal{H^+} &= \{ x \,|\, c^Tx \geq \cost_{\C}(c) \} \\
    \mathcal{H^-} &= \{ x \,|\, c^Tx \leq \support_{\D}(c) \}
\end{align}
From Lemma \ref{lem:sd_halfplane} $\sd (\mathcal{H}^+, \mathcal{H}^-) = \cost_{\C}(c) - \support_{\D}(c)$.
Noting that $\C \subset \mathcal{H}^+$ and $\D \subset \mathcal{H}^-$ yields the stated inequality. 

\end{proof}

\subsection{Proof of Lemma \ref{lem:sd_c_exists}}
We prove this for the case in which $\sd(\C,\D) < 0$. The $\sd(\C,\D) \geq 0$ case can be shown using similar arguments.

$\sd(\C,\D) = \gamma < 0 \implies \pen(\C,\D) = |\gamma| \implies \exists \, t \in \mathbb{R}^n, \|t\| = |\gamma|$ such that $(\C + t) \cap (\intD) = \emptyset$. As these are disjoint convex sets there exists a separating hyperplane $\cost_{\C}(c) \geq \support_{\intD}(c)$ for some $c \in \mathbb{R}^n \setminus 0$. By the scaling properties of the cost and support function we can take $\|c\| = 1$ w.l.o.g. Noting that $\support_{\intD}(c) = \support_{\D}(c)$ we obtain $\cost_{\C}(c) + c^Tt \geq \support_{\D}(c)$ for some $\|c\| = 1$. 
Assume $c \not= \frac{t}{\|t\|} \implies c^Tt < \|t\|$. 
Let $\kappa = (c^Tt) \implies \cost_\C(c) + c^T(\kappa c) \geq \support_\D(c) \implies \cost_{\C + \kappa c}(c) \geq \support_\D(c) \implies (\C + \kappa c) \cap (\intD) = \emptyset \implies \pen(\C,\D) \leq \kappa < \|t\| = |\gamma|$ a contradiction. Thus $c = \frac{t}{\|t\|}$.
Assume $\cost_\C(c) + c^Tt > \support_\D(c) \implies \cost_\C(c) + |\gamma| > \support_\D(c) \implies \sd(\C,\D) > -|\gamma|$ by Lemma \ref{lem:sd_cost_minus_support} a contradiction. Thus $\cost_{\C}(c) + c^Tt = \support_{\D}(c)$ for some $\|c\| = \frac{t}{\|t\|}$ where $(\C + t) \cap (\intD) = \emptyset$. 

\subsection{Proof of Lemma \ref{lem:sd_ball}}

\begin{proof}
From Lemma \ref{lem:sd_c_exists} there exists $c \in \mathbb{R}^n, \|c\| = 1$ such that $\cost_\C(c) - \support_\D(c) = \sd(\C,\D) \implies
\cost_\C(c) - r_\C\sqrt{c^Tc} - \support_\D(c) - r_\D\sqrt{c^Tc} = \sd(\C,\D) - r_\C - r_\D \implies 
\cost_{\C \oplus B_{r_\C}}(c) - \support_{\D \oplus B_{r_\D}}(c) = \sd(\C,\D) - r_\C - r_\D$.
From Lemma \ref{lem:sd_cost_minus_support}, this implies $\sd(\C \oplus B_{r_\C}, \D \oplus B_{r_\D}) \geq \sd(\C,\D) - r_\C - r_\D$. Assume $\sd(\C \oplus B_{r_\C}, \D \oplus B_{r_\D}) > \sd(\C,\D) - r_\C - r_\D \implies \exists \, c \in \mathbb{R}^n, \|c\| = 1$ such that 
$\cost_{\C \oplus B_{r_\C}}(c) - \support_{\D \oplus B_{r_\D}}(c) > \sd(\C,\D) - r_\C - r_\D
\implies 
\cost_\C(c) - r_\C\sqrt{c^Tc} - \support_\D(c) - r_\D\sqrt{c^Tc} > \sd(\C,\D) - r_\C - r_\D
\implies
\cost_\C(c) - \support_\D(c) > \sd(\C,\D)
\implies
\sd(\C,\D) > \sd(\C,\D)$ a contradiction. Thus
$\sd(\C \oplus B_{r_\C},\D \oplus B_{r_\D}) = \sd(\C,\D) - r_\C - r_\D$.
\end{proof}


\bibliographystyle{ieeetr}
\bibliography{myref}

\begin{thebibliography}{10}

\bibitem{Bock1984}
H.~Bock and K.~Plitt, ``A multiple shooting algorithm for direct solution of
  optimal control problems*,'' {\em IFAC Proceedings Volumes}, vol.~17, no.~2,
  pp.~1603--1608, 1984.

\bibitem{Biegler1984}
L.~T. Biegler, ``Solution of dynamic optimization problems by successive
  quadratic programming and orthogonal collocation,'' {\em Computers \&
  Chemical Engineering}, vol.~8, pp.~243--247, 1984.

\bibitem{Yan2015}
Y.~Yan and G.~S. Chirikjian, ``Closed-form characterization of the minkowski
  sum and difference of two ellipsoids,'' {\em Geometriae Dedicata}, vol.~177,
  no.~1, pp.~103--128, 2015.

\bibitem{Guthrie2022}
J.~Guthrie, M.~Kobilarov, and E.~Mallada, ``Closed-form minkowski sum
  approximations for efficient optimization-based collision avoidance,'' in
  {\em 2022 American Control Conference (ACC)}, pp.~3857--3864, 2022.

\bibitem{Patel2011}
R.~B. Patel and P.~J. Goulart, ``Trajectory generation for aircraft avoidance
  maneuvers using online optimization,'' {\em Journal of Guidance, Control, and
  Dynamics}, vol.~34, no.~1, pp.~218--230, 2011.

\bibitem{Gerdts2012}
M.~Gerdts, R.~Henrion, D.~Hömberg, and C.~Landry, ``Path planning and
  collision avoidance for robots,'' {\em Numerical Algebra, Control and
  Optimization}, vol.~2, no.~3, pp.~437--463, 2012.

\bibitem{Zhang2021}
X.~Zhang, A.~Liniger, and F.~Borrelli, ``Optimization-based collision
  avoidance,'' {\em IEEE Transactions on Control Systems Technology}, vol.~29,
  no.~3, pp.~972--983, 2021.

\bibitem{Boyd2004}
S.~Boyd and L.~Vandenberghe, {\em Convex Optimization}.
\newblock {Cambridge University Press}, March 2004.

\bibitem{Choi2009}
Y.-K. Choi, J.-W. Chang, W.~Wang, M.-S. Kim, and G.~Elber, ``Continuous
  collision detection for ellipsoids,'' {\em IEEE Transactions on Visualization
  and Computer Graphics}, vol.~15, no.~2, pp.~311--325, 2009.

\bibitem{Dueri2017}
D.~Dueri, Y.~Mao, Z.~Mian, J.~Ding, and B.~Açikmeşe, ``Trajectory
  optimization with inter-sample obstacle avoidance via successive
  convexification,'' in {\em 2017 IEEE 56th Annual Conference on Decision and
  Control (CDC)}, pp.~1150--1156, 2017.

\bibitem{Schulman2014}
J.~Schulman, Y.~Duan, J.~Ho, A.~X. Lee, I.~Awwal, H.~Bradlow, J.~Pan, S.~Patil,
  K.~Goldberg, and P.~Abbeel, ``Motion planning with sequential convex
  optimization and convex collision checking,'' {\em The International Journal
  of Robotics Research}, vol.~33, pp.~1251 -- 1270, 2014.

\bibitem{LaValle2006}
S.~M. LaValle, {\em Planning Algorithms}.
\newblock Cambridge University Press, 2006.

\bibitem{Schneider2013}
R.~Schneider, {\em Convex Bodies: The Brunn–Minkowski Theory}.
\newblock Encyclopedia of Mathematics and its Applications, Cambridge
  University Press, 2~ed., 2013.

\bibitem{Parillo2000}
P.~Parrilo, {\em Structured semidefinite programs and semialgebraic geometry
  methods in robustness and optimization}.
\newblock PhD thesis, California Institute of Technology, 2000.

\bibitem{Wachter2006}
A.~Wächter and L.~T. Biegler, ``On the implementation of an interior-point
  filter line-search algorithm for large-scale nonlinear programming,'' {\em
  Mathematical Programming}, vol.~106, no.~1, pp.~25--57, 2006.

\bibitem{Andersson2019}
J.~A.~E. Andersson, J.~Gillis, G.~Horn, J.~B. Rawlings, and M.~Diehl,
  ``{CasADi} -- {A} software framework for nonlinear optimization and optimal
  control,'' {\em Mathematical Programming Computation}, vol.~11, no.~1,
  pp.~1--36, 2019.

\end{thebibliography}

\end{document}